\documentclass[11pt,a4paper]{llncs}

\usepackage{latexsym}
\usepackage{amssymb}
\usepackage{amsmath}
\usepackage{enumerate}

\newcommand{\hitground}{\operatorname{\sf ground}}
\newcommand{\hitdom}{\operatorname{\sf dom}}
\newcommand{\hitlfp}{\operatorname{\sf lfp}}
\newcommand{\hitgfp}{\operatorname{\sf gfp}}
\newcommand{\hitGL}{\operatorname{GL}}
\newcommand{\hitCW}{\operatorname{CW}}
\newcommand{\hitCGL}{\operatorname{CGL}}

\begin{document}

\pagestyle{plain}

\mainmatter

\title{Towards a Systematic Account of Different Semantics for Logic
Programs}

\titlerunning{Towards a Systematic Account of Different Semantics for Logic
Programs}

\author{Pascal Hitzler\protect\footnote{The author acknowledges support by
the German Federal Ministry for Education and Research (BMBF) under the
SmartWeb project, and by the European Union under the KnowledgeWeb Network
of Excellence.}}

\authorrunning{Pascal Hitzler}

\institute{AIFB, Universit\"at Karlsruhe, Germany\\
http://www.aifb.uni-karlsruhe.de/WBS/phi/\\
hitzler@aifb.uni-karlsruhe.de}

\maketitle

\begin{abstract}
In \cite{HW02,HW05}, a new methodology has been proposed which allows to
derive uniform characterizations of different declarative semantics for
logic programs with negation. One result from this work is that the
well-founded semantics can formally be understood as a stratified version of
the Fitting (or Kripke-Kleene) semantics. The constructions leading to this
result, however, show a certain asymmetry which is not readily
understood. We will study this situation here with the result that we will
obtain a coherent picture of relations between different semantics for
normal logic programs.
\end{abstract}

\thispagestyle{plain}

\section{Introduction}

Within the past twenty years, many different declarative semantics for logic
programs with negation have been developed. Different perspectives on the
question what properties a semantics should foremost satisfy, have led to a
variety of diverse proposals. From a knowledge representation and reasoning
point of view it appears to be important that a semantics captures
established non-monotonic reasoning frameworks, e.g. Reiters default logic
\cite{Rei80}, and that they allow intuitively appealing, i.e. ``common
sense'', encodings of AI problems. The semantics which, due to common opinion
by researchers in the field, satisfy these requirements best, are the least
model semantics for definite programs \cite{Llo88}, and for normal programs
the stable \cite{GL88} and the well-founded semantics \cite{GRS91}. Of
lesser importance, albeit still acknowledged in particular for their
relation to resolution-based logic programming, are the Fitting semantics
\cite{Fit85} and approaches based on stratification \cite{ABW88,Prz88}. 

The semantics just mentioned are closely connnected by a number of well-
(and some lesser-) known relationships, and many authors have contributed to
this understanding. Fitting \cite{FitTCS} provides a framework using
Belnap's four-valued logic which encompasses supported, stable, Fitting, and
well-founded semantics. His work was recently extended by Denecker, Marek,
and Truszczynski \cite{DMT00}. Przymusinski \cite{Prz89} gives a version in
three-valued logic of the stable semantics, and shows that it coincides with
the well-founded one. Van Gelder \cite{Gel89} constructs the well-founded
semantics unsing the Gelfond-Lifschitz operator originally associated with
the stable semantics. Dung and Kanchanasut \cite{DK89} define the notion of
fixpoint completion of a program which provides connections between the
supported and the stable semantics, as well as between the Fitting and the
well-founded semantics, studied by Fages \cite{Fag94} and Wendt
\cite{Wen02jeeec}. Hitzler and Wendt \cite{HW02,HW05} have recently provided a
unifying framework using level mappings, and results which amongst other
things give further support to the point of view that the stable semantics
is a formal and natural extension to normal programs of the least model
semantics for definite programs. Furthermore, it was shown that the
well-founded semantics can be understood, formally, as a stratified version
of the Fitting semantics.

This latter result, however, exposes a certain asymmetry in the construction
leading to it, and it is natural to ask the question as to what exactly is
underlying it. This is what we will study in the sequel. In a nutshell, we
will see that formally this asymmetry is due to the well-known preference of
falsehood in logic programming semantics. More importantly, we will also see
that a ``dual'' theory, obtained from prefering truth, can be stablished
which is in perfect analogy to the close and well-known relationships
between the different semantics mentioned above. 

We want to make it explicit from the start that we do not intend to provide
new semantics for practical purposes\footnote{Although there may be some
virtue to this perspective, see \cite{Hit02circ}.}. We rather want to focus
on the deepening of the theoretical insights into the relations between
different semantics, by painting a coherent and complete picture of the
dependencies and interconnections. We find the richness of the theory very
appealing, and strongly supportive of the opinion that the major semantics
studied in the field are founded on a sound theoretical base. Indeed, from a
mathematical perspective one expects major notions in a field to be strongly
interconnected, and historic developments show that such foundational
underpinnings are supportive of a wide and lasting impact of a field. The
results in this paper aim at establishing these foundations in a clean and
formally satisfying manner.

The plan of the paper is as follows. In Section \ref{sec:prelim} we will
introduce notation and terminology needed for proving the results in the
main body of the paper. We will also review in detail those results from
\cite{HW02,HW05} which triggered and motivated our investigations. In
Section \ref{sec:proposal} we will provide a variant of the stable semantics
which prefers truth, and in Section \ref{sec:mcwf} we will do likewise for
the well-founded semantics. Throughout, our definitions will be accompanied
by results which complete the picture of relationships between different
semantics. 

This paper is a revised version of the conference contribution
\cite{Hit03ki}.

\bigskip

\noindent
\emph{Acknowledgements.} I am grateful for comments by anonymous referees
which helped to improve the presentation, and in  particular for bringing my
attention to the related and independent work reported in \cite{DBM01,DT04lpnmr}. 

\section{Preliminaries and Notation}\label{sec:prelim}

A (\emph{normal}) \emph{logic program} is a finite set of (universally
quantified) \emph{clauses} of the form $\forall(A\gets A_1\wedge\dots\wedge
A_n\wedge\lnot B_1\wedge\dots\wedge\lnot B_m)$, commonly written as $A\gets
A_1,\dots,A_n,\lnot B_1,\dots,\lnot B_m$, where $A$, $A_i$, and $B_j$, for
$i=1,\dots,n$ and $j=1,\dots,m$, are atoms over some given first order
language. $A$ is called the \emph{head} of the clause, while the remaining
atoms make up the \emph{body} of the clause, and depending on context, a
body of a clause will be a set of literals (i.e. atoms or negated atoms) or
the conjunction of these literals. Care will be taken that this
identification does not cause confusion. We allow a body, i.e. a
conjunction, to be empty, in which case it always evaluates to true. A
clause with empty body is called a \emph{unit clause} or a \emph{fact}. A
clause is called \emph{definite}, if it contains no negation symbol. A
program is called \emph{definite} if it consists only of definite
clauses. We will usually denote atoms with $A$ or $B$, and literals, which
may be atoms or negated atoms, by $L$ or $K$.

Given a logic program $P$, we can extract from it the components of a first
order language, and we always make the mild assumption that this language
contains at least one constant symbol. The corresponding set of ground
atoms, i.e. the \emph{Herbrand base} of the program, will be denoted by
$B_P$. For a subset $I\subseteq B_P$, we set $\lnot I=\{\lnot A\mid A\in
B_P\}$. The set of all ground instances of $P$ with respect to $B_P$ will be
denoted by $\hitground(P)$.  For $I\subseteq B_P\cup\lnot B_P$, we say that $A$
is \emph{true with respect to} (or \emph{in}) $I$ if $A\in I$, we say that
$A$ is \emph{false with respect to} (or \emph{in}) $I$ if $\lnot A\in I$,
and if neither is the case, we say that $A$ is \emph{undefined with respect
to} (or \emph{in}) $I$. A (\emph{three-valued} or \emph{partial})
\emph{interpretation} $I$ for $P$ is a subset of $B_P\cup\lnot B_P$ which is
\emph{consistent}, i.e. whenever $A\in I$ then $\lnot A\not\in I$. A body,
i.e. a conjunction of literals, is true in an interpretation $I$ if every
literal in the body is true in $I$, it is false in $I$ if one of its
literals is false in $I$, and otherwise it is undefined in $I$. For a
negated literal $L=\lnot A$ we will find it convenient to write $\lnot L\in
I$ if $A\in I$. By $I_P$ we denote the set of all (three-valued)
interpretations of $P$. Both $I_P$ and $B_P\cup\lnot B_P$ are complete
partial orders (cpos) via set-inclusion, i.e. they contain the empty set as
least element, and every ascending chain has a supremum, namely its union. A
\emph{model} of $P$ is an interpretation $I\in I_P$ such that for each
clause $A\gets\texttt{body}$ we have that $\texttt{body}\subseteq I$ implies $A\in I$.  A
\emph{total} interpretation is an interpretation $I$ such that no $A\in B_P$
is undefined in $I$.

For an interpretation $I$ and a program $P$, an \emph{$I$-partial level
mapping} for $P$ is a partial mapping $l:B_P\to\alpha$ with domain
$\hitdom(l)=\{A\mid A\in I\text{ or }\lnot A\in I\}$, where $\alpha$ is some
(countable) ordinal. We extend every level mapping to literals by setting
$l(\lnot A)=l(A)$ for all $A\in\hitdom(l)$. A (\emph{total}) \emph{level
mapping} is a total mapping $l:B_P\to\alpha$ for some (countable) ordinal
$\alpha$.

Given a normal logic program $P$ and some $I\subseteq B_P\cup\lnot B_P$, we say
that $U\subseteq B_P$ is an \emph{unfounded set} (\emph{of $P$}) \emph{with
respect to $I$} if each atom $A\in U$ satisfies the following condition: For
each clause $A\gets\texttt{body}$ in $\hitground(P)$ (at least) one of the following
holds.
\begin{enumerate}[(Ui)]
\item Some (positive or negative) literal in $\texttt{body}$ is false in $I$.
\item Some (non-negated) atom in $\texttt{body}$ occurs in $U$.
\end{enumerate}

Given a normal logic program $P$, we define the following operators on
$B_P\cup\lnot B_P$. $T_P(I)$ is the set of all $A\in B_P$ such that there
exists a clause $A\gets\texttt{body}$ in $\hitground(P)$ such that $\texttt{body}$ is true in
$I$. $F_P(I)$ is the set of all $A\in B_P$ such that for all clauses
$A\gets\texttt{body}$ in $\hitground(P)$ we have that $\texttt{body}$ is false in $I$. Both
$T_P$ and $F_P$ map elements of $I_P$ to elements of $I_P$. Now define the
operator $\Phi_P: I_P\to I_P$ by
$$
\Phi_P(I) = T_P(I)\cup\lnot F_P(I).
$$ This operator is due to \cite{Fit85} and is well-defined and monotonic on
the cpo $I_P$, hence has a least fixed point by the
Knaster-Tarski\footnote{\newcounter{fptfootnote}%
\setcounter{fptfootnote}{\thefootnote}%
We follow the terminology from \cite{Jac01}. The Knaster-Tarski theorem is
sometimes called Tarski theorem and states that every monotonic function on
a cpo has a least fixed point, which can be obtained by transfinitely
iterating the bottom element of the cpo. The Tarski-Kantorovitch theorem is
sometimes refered to as the Kleene theorem or the Scott theorem (or even as
``the'' fixed-point theorem) and states that if the function is additionally
Scott (or order-) continuous, then the least fixed point can be obtained by
an iteration which is not transfinite, i.e. closes off at $\omega$, the least
infinite ordinal. In both cases, the least fixed point is also the least
pre-fixed point of the function.}  fixed-point theorem, and we can obtain
this fixed point by defining, for each monotonic operator $F$, that $F\!\uparrow\!
0=\emptyset$, $F\!\uparrow\!(\alpha+1)=F(F\!\uparrow\!\alpha)$ for any ordinal $\alpha$, and
$F\!\uparrow\!\beta=\bigcup_{\gamma<\beta}F\!\uparrow\!\gamma$ for any limit ordinal
$\beta$, and the least fixed point of $F$ is obtained as $F\!\uparrow\!\alpha$ for
some ordinal $\alpha$. The least fixed point of $\Phi_P$ is called the
\emph{Kripke-Kleene model} or \emph{Fitting model} of $P$, determining the
\emph{Fitting semantics} of $P$.

Now, for $I\subseteq B_P\cup\lnot B_P$, let $U_P(I)$ be the greatest unfounded set
(of $P$) with respect to $I$, which always exists due to
\cite{GRS91}. Finally, define 
$$
W_P(I) = T_P(I)\cup\lnot U_P(I)
$$ for all $I\subseteq B_P\cup\lnot B_P$. The operator $W_P$, which operates
on the cpo $B_P\cup\lnot B_P$, is due to \cite{GRS91} and is monotonic,
hence has a least fixed point by the Knaster-Tarski\newcounter{tempfootnote}%
\setcounter{tempfootnote}{\thefootnote}%
\setcounter{footnote}{\thefptfootnote}%
\addtocounter{footnote}{-1}%
\footnotemark%
\setcounter{footnote}{\thetempfootnote}{} fixed-point theorem, as
above for $\Phi_P$. It turns out that $W_P\!\uparrow\!\alpha$ is in $I_P$ for each
ordinal $\alpha$, and so the least fixed point of $W_P$ is also in $I_P$ and
is called the \emph{well-founded model} of $P$, giving the
\emph{well-founded semantics} of $P$.

In order to avoid confusion, we will use the following terminology: the
notion of \emph{interpretation}, and $I_P$ will be the set of all those,
will by default denote consistent subsets of $B_P\cup\lnot B_P$,
i.e. interpretations in three-valued logic. We will sometimes emphasize this
point by using the notion \emph{partial interpretation}. By \emph{two-valued
interpretations} we mean subsets of $B_P$. Both interpretations and
two-valued interpretations are ordered by subset inclusion. Each two-valued
interpretation $I$ can be identified with the partial interpretation
$I'=I\cup\lnot (B_P\setminus I)$. Note however, that in this case $I'$ is
always a maximal element in the ordering for partial interpretations, while
$I$ is in general not maximal as a two-valued interpretation\footnote{These
two orderings in fact correspond to the knowledge and truth orderings as
discussed in \cite{Fit91}.}. Given a partial interpretation $I$, we set
$I^+=I\cap B_P$ and $I^-= \{A\in B_P\mid \lnot A\in I\}$.

Given a program $P$, we define the operator $T_P^+$ on subsets of $B_P$ by
$T_P^+(I)=T_P(I\cup\lnot (B_P\setminus I))$. The pre-fixed points of
$T_P^+$, i.e. the two-valued interpretations $I\subseteq B_P$ with
$T_P^+(I)\subseteq I$, are exactly the models, in the sense of classical
logic, of $P$. Post-fixed points of $T_P^+$, i.e. $I\subseteq B_P$ with
$I\subseteq T_P^+(I)$ are called \emph{supported interpretations} of $P$,
and a supported model of $P$ is a model $P$ which is a supported
interpretation. The supported models of $P$ thus coincide with the fixed
points of $T_P^+$. It is well-known that for definite programs $P$ the
operator $T_P^+$ is monotonic on the set of all subsets of $B_P$, with
respect to subset inclusion. Indeed it is Scott-continuous
\cite{Llo88,AJ94} and, via the Tarski-Kantorovich%
\newcounter{tmpfootnote}%
\setcounter{tmpfootnote}{\thefootnote}%
\setcounter{footnote}{\thefptfootnote}%
\addtocounter{footnote}{-1}%
\footnotemark%
\setcounter{footnote}{\thetmpfootnote}{} fixed-point theorem, achieves its
least pre-fixed point $M$, which is also a fixed point, as the supremum of
the iterates $T_P^+\!\uparrow\! n$ for $n\in\mathbb{N}$. So
$M=\hitlfp\left(T_P^+\right)=T_P^+\!\uparrow\!\omega$ is \emph{the least two-valued
model} of $P$. Likewise, since the set of all subsets of $B_P$ is a complete
lattice, and therefore has greatest element $B_P$, we can also define
$T_P^+\!\downarrow\! 0=B_P$ and inductively $T_P^+\!\downarrow\!
(\alpha+1)=T_P^+(T_P^+\!\downarrow\!\alpha)$ for each ordinal $\alpha$ and
$T_P^+\!\downarrow\!\beta=\bigcap_{\gamma<\beta} T_P^+\!\downarrow\!\gamma$ for each limit
ordinal $\beta$. Again by the Knaster-Tarski fixed-point theorem, applied to
the superset inclusion ordering (i.e. reverse subset inclusion) on subsets
of $B_P$, it turns out that $T_P^+$ has a greatest fixed point,
$\hitgfp\left(T_P^+\right)$.

The stable model semantics due to \cite{GL88} is intimately related to the
well-founded semantics. Let $P$ be a normal program, and let $M\subseteq
B_P$ be a set of atoms. Then we define $P/M$ to be the (ground) program
consisting of all clauses $A\gets A_1,\dots,A_n$ for which there is a clause
$A\gets A_1,\dots,A_n,\lnot B_1,\dots,\lnot B_m$ in $\hitground(P)$ with
$B_1,\dots,B_m\not\in M$. Since $P/M$ does no longer contain negation, it
has a least two-valued model $T_{P/M}^+\!\uparrow\!\omega$. For any two-valued
interpretation $I$ we can therefore define the operator
$\hitGL_P(I)=T_{P/I}^+\!\uparrow\!\omega$, and call $M$ a \emph{stable model} of the
normal program $P$ if it is a fixed point of the operator $\hitGL_P$, i.e. if
$M=\hitGL_P(M)=T_{P/M}^+\!\uparrow\!\omega$. As it turns out, the operator $\hitGL_P$ is
in general not monotonic for normal programs $P$. However it is
\emph{antitonic}, i.e. whenever $I\subseteq J\subseteq B_P$ then
$\hitGL_P(J)\subseteq\hitGL_P(I)$. As a consequence, the operator $\hitGL_P^2$,
obtained by applying $\hitGL_P$ twice, is monotonic, and hence has a least
fixed point $L_P$ and a greatest fixed point $G_P$. In \cite{Gel89} it was
shown that $\hitGL_P(L_P)=G_P$, $L_P=\hitGL_P(G_P)$, and that $L_P\cup\lnot
(B_P\setminus G_P)$ coincides with the well-founded model of $P$. This is
called the \emph{alternating fixed point characterization} of the
well-founded semantics.

\subsection*{Some Results}

The following is a straightforward result which has, to the best of our
knowledge, first been formally reported in \cite{HW05}, where a proof
can be found.

\begin{theorem}\label{theo:defleast}
Let $P$ be a definite program. Then there is a unique two-valued model $M$
of $P$ for which there exists a (total) level mapping $l: B_P\to\alpha$ such
that for each atom $A\in M$ there exists a clause $A\gets A_1,\dots,A_n$ in
$\hitground(P)$ with $A_i\in M$ and $l(A)>l(A_i)$ for all
$i=1,\dots,n$. Furthermore, $M$ is the least two-valued model of $P$.
\end{theorem}

The following result is due to \cite{Fag94}, and is striking in its
similarity to Theorem~\ref{theo:defleast}. 

\begin{theorem}\label{theo:stablechar}
Let $P$ be normal. Then a two-valued model $M\subseteq B_P$ of $P$ is a
stable model of $P$ if and only if there exists a (total) level mapping
$l:B_P\to\alpha$ such that for each $A\in M$ there exists $A\gets
A_1,\dots,A_n\lnot B_1,\dots,\lnot B_m$ in $\hitground(P)$ with $A_i\in M$,
$B_j\not\in M$, and $l(A)>l(A_i)$ for all $i=1,\dots,n$ and $j=1,\dots,m$.
\end{theorem}

We next recall the following alternative characterization of the Fitting
model, due to \cite{HW02,HW05}.

\begin{definition}\label{def:fit}
Let $P$ be a normal logic program, $I$ be a model of $P$, and $l$ be an
$I$-partial level mapping for $P$. We say that $P$ \emph{satisfies} (F)
\emph{with respect to $I$ and $l$}, if each $A\in\hitdom(l)$ satisfies one of
the following conditions.
\begin{enumerate}[(Fi)]
\item $A\in I$ and there exists a clause $A\gets L_1,\dots, L_n$ in
$\hitground(P)$ such that $L_i\in I$ and $l(A)>l(L_i)$ for all
$i$.
\item $\lnot A\in I$ and for each clause $A\gets L_1,\dots, L_n$ in
$\hitground(P)$ there exists $i$ with $\lnot L_i\in I$ and
$l(A)>l(L_i)$.
\end{enumerate}
\end{definition}

\begin{theorem}\label{theo:fitchar}
Let $P$ be a normal logic program with Fitting model $M$. Then $M$ is the
greatest model among all models $I$, for which there exists an $I$-partial
level mapping $l$ for $P$ such that $P$ satisfies (F) with respect to $I$
and $l$.
\end{theorem}

Let us recall next the definition of a (locally) stratified program, due to
\cite{ABW88,Prz88}: A normal logic program is called \emph{locally
stratified} if there exists a (total) level mapping $l:B_P\to\alpha$, for
some ordinal $\alpha$, such that for each clause $A\gets A_1,\dots,A_n,
\lnot B_1,\dots,\lnot B_m$ in $\hitground(P)$ we have that $l(A)\geq l(A_i)$
and $l(A)>l(B_j)$ for all $i=1,\dots,n$ and $j=1,\dots,m$.  The notion of
(locally) stratifed program was developed with the idea of preventing
\emph{recursion through negation}, while allowing recursion through positive
dependencies. (Locally) stratified programs have total well-founded models.

There exist locally stratified programs which do not have a total Fitting
semantics and vice versa --- just consider the programs consisting of the
single clauses $p\gets p$, respectively, $p\gets\lnot p,q$. In fact,
condition (Fii) requires a strict decrease of level between the head and a
literal in the rule, independent of this literal being positive or
negative. But, on the other hand, condition (Fii) imposes no further
restrictions on the remaining body literals, while the notion of local
stratification does. These considerations motivate the substitution of
condition (Fii) by the condition (Cii), as done for the following
definition.

\begin{definition}\label{def:wfchar}
Let $P$ be a normal logic program, $I$ be a model of $P$, and $l$ be an
$I$-partial level mapping for $P$. We say that \emph{$P$ satisfies} (WF)
\emph{with respect to $I$ and $l$}, if each $A\in\hitdom(l)$ satisfies (Fi) or
the following condition.
\begin{enumerate}[(Cii)]
\item[(Cii)] $\lnot A\in I$ and for each clause $A\gets A_1,\dots, A_n,\lnot
B_1,\dots,\lnot B_m$ contained in $\hitground(P)$ (at least) one of the
following conditions holds:
\begin{enumerate}[({Cii}a)]
\item There exists $i\in\{1,\dots,n\}$ with $\lnot A_i\in I$ and
$l(A)\geq l(A_i)$.
\item There exists $j\in\{1,\dots,m\}$ with $B_j\in I$ and $l(A)>l(B_j)$. 
\end{enumerate}
\end{enumerate}
\end{definition}

So, in the light of Theorem \ref{theo:fitchar}, Definition \ref{def:wfchar}
should provide a natural ``stratified version'' of the Fitting
semantics. And indeed it does, and furthermore, the resulting semantics
coincides with the well-founded semantics, which is a very satisfactory
result from \cite{HW02,HW05}. 

\begin{theorem}\label{theo:wfchar}
Let $P$ be a normal logic program with well-founded model $M$. Then $M$ is
the greatest model among all models $I$, for which there exists an
$I$-partial level mapping $l$ for $P$ such that $P$ satisfies (WF) with
respect to $I$ and $l$.
\end{theorem}

For completeness, we remark that an alternative characterization of the
weakly perfect model semantics \cite{PP90} can also be found in
\cite{HW02,HW05}. 

The approach which led to the results just mentioned, originally put forward
in \cite{HW02,HW05}, provides a general methodology for obtaining
uniform characterizations of different semantics for logic programs.

\section{Maximally Circular Stable Semantics}\label{sec:proposal}

We note that condition (Fi) has been reused in Definition
\ref{def:wfchar}. Thus, Definition \ref{def:fit} has been ``stratified''
only with respect to condition (Fii), yielding (Cii), but not with respect
to (Fi). Indeed, also replacing (Fi) by a stratified version such as the
following seems not satisfactory at first sight.
\begin{enumerate}[(Ci)]
\item $A\in I$ and there exists a clause $A\gets A_1,\dots, A_n,\lnot
B_1,\dots,\lnot B_m$ in $\hitground(P)$ such that $A_i,\lnot B_j\in I$,
$l(A)\geq l(A_i)$, and $l(A)>l(B_j)$ for all $i$ and $j$.
\end{enumerate}
If we replace condition (Fi) by condition (Ci) in Definition
\ref{def:wfchar}, then it is not guaranteed that for any given program there
is a greatest model satisfying the desired properties, as the following
example from \cite{HW02,HW05} shows.

\begin{example}\label{bsp:oandwf} 
Consider the program consisting of the two clauses $p\gets p$ and
$q\gets\lnot p$, and the two (total) models $M_1=\{p,\lnot q\}$ and
$M_2=\{\lnot p,q\}$, which are incomparable, and the level mapping $l$ with
$l(p)=0$ and $l(q)=1$.
\end{example}

In order to arrive at an understanding of this asymmetry, we consider the
setting with conditions (Ci) and (Fii), which is somehow ``dual'' to the
well-founded semantics which is characterized by (Fi) and (Cii).

\begin{definition}\label{def:owfchar}
Let $P$ be a normal logic program, $I$ be a model of $P$, and $l$ be an
$I$-partial level mapping for $P$. We say that \emph{$P$ satisfies} (CW)
\emph{with respect to $I$ and $l$}, if each $A\in\hitdom(l)$ satisfies (Ci) or
(Fii).
\end{definition}

By virtue of Definition \ref{def:owfchar} we will be able to develop a
theory which complements the restults from Section \ref{sec:prelim}. We will
first characterize the greatest model of a definite program analogously to
Theorem \ref{theo:defleast}.

\begin{theorem}\label{theo:defgreatest}
Let $P$ be a definite program. Then there is a unique two-valued supported
interpretation $M$ of $P$ for which there exists a (total) level mapping $l:
B_P\to\alpha$ such that for each atom $A\not\in M$ and for all clauses
$A\gets A_1,\dots,A_n$ in $\hitground(P)$ there is some $A_i\not\in M$ with
$l(A)>l(A_i)$. Furthermore, $M$ is the greatest two-valued model of $P$.
\end{theorem}

\begin{proof}
Let $M$ be the greatest two-valued model of $P$, and let $\alpha$ be the
least ordinal such that $M=T_P^+\!\downarrow\!\alpha$. Define $l:B_P\to\alpha$
by setting $l(A)=\min\{\gamma\mid A\not\in T_P^+\!\downarrow\!(\gamma+1)\}$ for
$A\not\in M$, and by setting $l(A)=0$ if $A\in M$. The mapping $l$ is
well-defined because $A\not\in M$ with $A\not\in
T_P^+\!\downarrow\!\gamma=\bigcap_{\beta<\gamma}T_P^+\!\downarrow\!\beta$ for some limit
ordinal $\gamma$ implies $A\not\in T_P^+\!\downarrow\!\beta$ for some
$\beta<\gamma$. So the least ordinal $\beta$ with $A\not\in
T_P^+\!\downarrow\!\beta$ is always a successor ordinal. Now assume that there is
$A\not\in M$ which does not satisfy the stated condition. We can furthermore
assume without loss of generality that $A$ is chosen with this property such
that $l(A)$ is minimal. Let $A\gets A_1,\dots,A_n$ be a clause in
$\hitground(P)$. Since $A\not\in T_P^+\left(T_P^+\!\downarrow\! l(A)\right)$ we obtain
$A_i\not\in T_P^+\!\downarrow\! l(A)\supseteq M$ for some $i$. But then
$l(A_i)<l(A)$ which contradicts minimality of $l(A)$.

Conversely, let $M$ be a two-valued model for $P$ which satisfies the given
condition for some mapping $l:B_P\to\alpha$. We show by transfinite
induction on $l(A)$ that $A\not\in M$ implies $A\not\in T_P^+\!\downarrow\!
(l(A)+1)$, which suffices because it implies that for the greatest
two-valued model $T_P^+\!\downarrow\!\beta$ of $P$ we have that
$T_P^+\!\downarrow\!\beta\subseteq M$, and therefore $T_P^+\!\downarrow\!\beta=M$. For the
inductive proof consider first the case where $l(A)=0$. Then there is no
clause in $\hitground(P)$ with head $A$ and consequently $A\not\in T_P^+\!\downarrow\!
1=T_P^+(B_P)$. Now assume that the statement to be proven holds for all
$B\not\in M$ with $l(B)<\alpha$, where $\alpha$ is some ordinal, and let
$A\not\in M$ with $l(A)=\alpha$. Then each clause in $\hitground(P)$ with head
$A$ contains an atom $B$ with $l(B)=\beta<\alpha$ and $B\not\in M$. Hence
$B\not\in T_P^+\!\downarrow\!(\beta+1)$ and consequently $A\not\in
T_P^+\!\downarrow\!(\alpha+1)$. 
\end{proof}

The following definition and theorem are analogous to Theorem
\ref{theo:stablechar}. 

\begin{definition}\label{def:ostable}
Let $P$ be normal. Then $M\subseteq B_P$ is called a \emph{maximally
circular stable model} (\emph{maxstable model}) of $P$ if it is a two-valued
supported interpretation of $P$ and there exists a (total) level mapping
$l:B_P\to\alpha$ such that for each atom $A\not\in M$ and for all clauses
$A\gets A_1,\dots,A_n,\lnot B_1,\dots,\lnot B_m$ in $\hitground(P)$ with
$B_1,\dots,B_m\not\in M$ there is some $A_i\not\in M$ with $l(A)>l(A_i)$.
\end{definition}

\begin{theorem}\label{theo:ostable}
$M\subseteq B_P$ is a maxstable model of $P$ if and only if
$M=\hitgfp\left(T_{P/M}^+\right)$.
\end{theorem}

\begin{proof}
First note that every maxstable model is a a supported model. Indeed
supportedness follows immediately from the definition. Now assume that $M$
is maxstable but is not a model, i.e. there is $A\not\in M$ but there is a
clause $A\gets A_1,\dots,A_n$ in $\hitground(P)$ with $A_i\in M$ for all
$i$. But by the definition of maxstable model we must have that there is
$A_i\not\in M$, which contradicts $A_i\in M$.

Now let $M$ be a maxstable model of $P$. Let $A\not\in M$ and let
$T_{P/M}^+\!\downarrow\!\alpha=\hitgfp\left(T_{P/M}^+\right)$. We show by transfinite
induction on $l(A)$ that $A\not\in T_{P/M}^+\!\downarrow\! (l(A)+1)$ and hence
$A\not\in T_{P/M}^+\!\downarrow\!\alpha$. For $l(A)=0$ there is no clause with head
$A$ in $P/M$, so $A\not\in T_{P/M}^+\!\downarrow\! 1$. Now let $l(A)=\beta$ for some
ordinal $\beta$. By assumption we have that for all clauses $A\gets
A_1,\dots,A_n,\lnot B_1,\dots,\lnot B_m$ with $B_1,\dots,B_m\not\in M$ there
exists $A_i\not\in M$ with $l(A)>l(A_i)$, say $l(A_i)=\gamma<\beta$. Hence
$A_i\not\in T_{P/M}^+\!\downarrow\!(\gamma+1)$, and consequently $A\not\in
T_{P/M}^+\!\downarrow\!(\beta+1)$, which shows that
$\hitgfp\left(T_{P/M}^+\right)\subseteq M$.

So let again $M$ be a maxstable model of $P$ and let
$A\not\in\hitgfp\left(T_{P/M}^+\right)=T_{P/M}^+\!\downarrow\!\alpha$ and
$l(A)=\beta$. Then for each clause $A\gets A_1,\dots, A_n$ in $P/M$ there is
$A_i$ with $A_i\not\in T_{P/M}^+\!\downarrow\!\alpha$ and $l(A)>l(A_i)$. Now assume
$A\in M$. Without loss of generality we can furthermore assume that $A$ is
chosen such that $l(A)=\beta$ is minimal. Hence $A_i\not\in M$, and we
obtain that for each clause in $P/M$ with head $A$ one of the corresponding
body atoms is false in $M$. By supportedness of $M$ this yields $A\not\in
M$, which contradicts our assumption. Hence $A\not\in M$ as desired.

Conversely, let $M=\hitgfp\left(T_{P/M}^+\right)$. Then as an immediate
consequence of Theorem \ref{theo:defgreatest} we obtain that $M$ is
maxstable. 
\end{proof}

\section{Maximally Circular Well-Founded Semantics}\label{sec:mcwf}

Maxstable models are formally analogous\footnote{The term \emph{dual} seems
not to be entirely adequate in this situation, although it is intuitionally
appealing.} to stable models in that the former are fixed points of the
operator $I\mapsto\hitgfp\left(T_{P/I}^+\right)$, while the latter are fixed
points of the operator $I\mapsto\hitlfp\left(T_{P/I}^+\right)$. Further, in
analogy to the alternating fixed point characterization of the well-founded
model, we can obtain a corresponding variant of the well-founded semantics,
which we will do next. Theorem \ref{theo:ostable} suggests the definition
of the following operator.

\begin{definition}\label{def:oglop}
Let $P$ be a normal program and $I$ be a two-valued interpretation. Then
define $\hitCGL_P(I)=\hitgfp\left(T_{P/I}^+\right)$.
\end{definition}

Using the operator $\hitCGL_P$, we can define a ``maximally circular'' version
of the alternating fixed-point semantics.

\begin{proposition}\label{prop:oafp}
Let $P$ be a normal program. Then the following hold.
\begin{enumerate}[(i)]
\item $\hitCGL_P$ is antitonic and $\hitCGL_P^2$ is monotonic.
\item $\hitCGL_P\left(\hitlfp\left(\hitCGL_P^2\right)\right)=\hitgfp\left(\hitCGL_P^2\right)$ and
$\hitCGL_P\left(\hitgfp\left(\hitCGL_P^2\right)\right)=\hitlfp\left(\hitCGL_P^2\right)$.
\end{enumerate}
\end{proposition}

\begin{proof}
(i) If $I\subseteq J\in B_P$, then $P/J\subseteq P/I$ and consequently
$\hitCGL_P(J)=\hitgfp\left(T_{P/J}^+\right)\subseteq
\hitgfp\left(T_{P/I}^+\right)=\hitCGL_P(I)$. Monotonicity of $\hitCGL_P^2$ then follows
trivially. 

(ii) Let $L_P=\hitlfp\left(\hitCGL_P^2\right)$ and
$G_P=\hitgfp\left(\hitCGL_P^2\right)$. Then we can calculate
$\hitCGL_P^2(\hitCGL_P(L_P)) = \hitCGL_P\left(\hitCGL_P^2(L_P)\right) = \hitCGL_P(L_P)$, so
$\hitCGL_P(L_P)$ is a fixed point of $\hitCGL_P^2$, and hence $L_P\subseteq
\hitCGL_P(L_P)\subseteq G_P$. Similarly, $L_P\subseteq \hitCGL_P(G_P)\subseteq
G_P$. Since $L_P\subseteq G_P$ we get from the antitonicity of $\hitCGL_P$ that
$L_P\subseteq \hitCGL_P(G_P)\subseteq \hitCGL_P(L_P)\subseteq G_P$. Similarly,
since $\hitCGL_P(L_P)\subseteq G_P$, we obtain $\hitCGL_P(G_P)\subseteq
\hitCGL_P^2(L_P)=L_P\subseteq \hitCGL_P(G_P)$, so $\hitCGL_P(G_P)=L_P$, and also
$G_P=\hitCGL_P^2(G_P)=\hitCGL_P(L_P)$.
\end{proof}

We will now define an operator for the maximally circular well-founded
semantics. Given a normal logic program $P$ and some $I\in I_P$, we say that
$S\subseteq B_P$ is a \emph{self-founded set} (\emph{of $P$}) \emph{with
respect to $I$} if $S\cup I\in I_P$ and each atom $A\in S$ satisfies the
following condition: There exists a clause $A\gets\texttt{body}$ in $\hitground(P)$
such that one of the following holds.
\begin{enumerate}[(Si)]
\item $\texttt{body}$ is true in $I$.
\item Some (non-negated) atoms in $\texttt{body}$ occur in $S$ and all other
literals in $\texttt{body}$ are true in $I$.
\end{enumerate}

Self-founded sets are analogous\footnote{Again, it is not really a duality.}
to unfounded sets, and the following proposition holds.

\begin{proposition}\label{prop:greatepifound}
Let $P$ be a normal program and let $I\in I_P$. Then there exists a greatest
self-founded set of $P$ with respect to $I$.
\end{proposition}

\begin{proof}
If $(S_i)_{i\in{\mathcal{I}}}$ is a family of sets each of which is a self-founded set
of $P$ with respect to $I$, then it is easy to see that $\bigcup_{i\in{\mathcal{I}}}
S_i$ is also a self-founded set of $P$ with respect to $I$.
\end{proof}

Given a normal program $P$ and $I\in I_P$, let $S_P(I)$ be the greatest
self-founded set of $P$ with respect to $I$, and define the operator $\hitCW_P$
on $I_P$ by
$$\hitCW_P(I)= S_P(I)\cup \lnot F_P(I).$$

\begin{proposition}\label{prop:owmono}
The operator $\hitCW_P$ is well-defined and monotonic.
\end{proposition}

\begin{proof}
For well-definedness, we have to show that $S_P(I)\cap F_P(I)=\emptyset$ for
all $I\in I_P$. So assume there is $A\in S_P(I)\cap F_P(I)$. From $A\in
F_P(I)$ we obtain that for each clause with head $A$ there is a
corresponding body literal $L$ which is false in $I$. From $A\in S_P(I)$,
more precisely from (Sii), we can furthermore conclude that $L$ is an atom
and $L\in S_P(I)$. But then $\lnot L\in I$ and $L\in S_P(I)$ which is
impossible by definition of self-founded set which requires that $S_P(I)\cup
I\in I_P$. So $S_P(I)\cap F_P(I)=\emptyset$ and $\hitCW_P$ is well-defined.

For monotonicity, let $I\subseteq J\in I_P$ and let $L\in \hitCW_P(I)$. If
$L=\lnot A$ is a negated atom, then $A\in F_P(I)$ and all clauses with head
$A$ contain a body literal which is false in $I$, hence in $J$, and we
obtain $A\in F_P(J)$. If $L=A$ is an atom, then $A\in S_P(I)$ and there
exists a clause $A\gets\texttt{body}$ in $\hitground(P)$ such that (at least) one of
(Si) or (Sii) holds. If (Si) holds, then $\texttt{body}$ is true in $I$, hence in
$J$, and $A\in S_P(J)$. If (Sii) holds, then some non-negated atoms in
$\texttt{body}$ occur in $S$ and all other literals in $\texttt{body}$ are true in $I$,
hence in $J$, and we obtain $A\in S_P(J)$.
\end{proof}

Since Proposition \ref{prop:owmono} establishes monotonicity of $\hitCW_P$,
for normal $P$, we conclude that this operator has a least fixed point
$\hitlfp(\hitCW_P)$. 

\begin{definition}\label{def:maxwf}
For a normal program $P$, we call $\hitlfp(\hitCW_P)$ the \emph{maximally
circular well-founded model} (\emph{maxwf model}) of $P$.
\end{definition}

The following theorem relates our observations to Definition
\ref{def:owfchar}, in perfect analogy to the correspondence between the
stable model semantics, Theorem \ref{theo:defleast}, Fages's
characterization from Theorem \ref{theo:stablechar}, the well-founded
semantics, and the alternating fixed point characterization. 

\begin{theorem}\label{theo:main}
Let $P$ be a normal program and $M_P=\hitlfp(\hitCW_P)$ be its maxwf
model. Then the following hold.
\begin{enumerate}[(i)]
\item $M_P$ is the greatest model among all models $I$ of $P$ such that there
is an $I$-partial level mapping $l$ for $P$ such that $P$ satisfies (CW) with
respect to $I$ and $l$. 
\item
$M_P=\hitlfp\left(\hitCGL_P^2\right)\cup\lnot\left(B_P\setminus\hitgfp\left(\hitCGL_P^2\right)\right)$. 
\end{enumerate}
\end{theorem}

\begin{proof}
(i) Let $M_P=\hitlfp(\hitCW_P)$ and define the $M_P$-partial level mapping $l_P$ as
follows: $l_P(A)=\alpha$, where $\alpha$ is the least ordinal such that $A$
is not undefined in $\hitCW_P\!\uparrow\!(\alpha+1)$. The proof will be established by
showing the following facts: (1) $P$ satisfies (CW) with respect to $M_P$
and $l_P$. (2) If $I$ is a model of $P$ and $l$ is an $I$-partial level
mapping such that $P$ satisfies (CW) with respect to $I$ and $l$, then
$I\subseteq M_P$.

(1) Let $A\in\hitdom(l_P)$ and $l_P(A)=\alpha$. We consider two cases.

(Case i) If $A\in M_P$, then $A\in S_P(\hitCW_P\!\uparrow\!\alpha)$, hence there exists
a clause $A\gets\texttt{body}$ in $\hitground(P)$ such that (Si) or (Sii) holds with
respect to $\hitCW_P\!\uparrow\!\alpha$. If (Si) holds, then all literals in $\texttt{body}$
are true in $\hitCW_P\!\uparrow\!\alpha$, hence have level less than $l_P(A)$ and (Ci)
is satisfied. If (Sii) holds, then some non-negated atoms from $\texttt{body}$ occur
in $S_P(\hitCW_P\!\uparrow\!\alpha)$, hence have level less than or equal to $l_P(A)$,
and all remaining literals in $\texttt{body}$ are true in $\hitCW_P\!\uparrow\!\alpha$, hence
have level less than $l_P(A)$. Consequently, $A$ satisfies (Ci) with respect
to $M_P$ and $l_P$.

(Case ii) If $\lnot A\in M_P$, then $A\in F_P(\hitCW_P\!\uparrow\!\alpha)$, hence for
all clauses $A\gets\texttt{body}$ in $\hitground(P)$ there exists $L\in\texttt{body}$ with
$\lnot L\in\hitCW_P\!\uparrow\!\alpha$ and $l_P(L)<\alpha$, hence $\lnot L\in
M_P$. Consequently, $A$ satisfies (Fii) with respect to $M_P$ and $l_P$,
and we have established that fact (1) holds.

(2) We show via transfinite induction on $\alpha=l(A)$, that whenever $A\in
I$ (respectively, $\lnot A\in I$), then $A\in\hitCW_P\!\uparrow\!(\alpha+1)$
(respectively, $\lnot A\in\hitCW_P\!\uparrow\!(\alpha+1)$). For the base case, note
that if $l(A)=0$, then $\lnot A\in I$ implies that there is no clause with
head $A$ in $\hitground(P)$, hence $\lnot A\in\hitCW_P\!\uparrow\! 1$. If $A\in I$ then
consider the set $S$ of all atoms $B$ with $l(B)=0$ and $B\in I$. We show
that $S$ is a self-founded set of $P$ with respect to $\hitCW_P\!\uparrow\!
0=\emptyset$, and this suffices since it implies $A\in\hitCW_P\!\uparrow\! 1$ by the
fact that $A\in S$. So let $C\in S$. Then $C\in I$ and $C$ satisfies
condition (Ci) with respect to $I$ and $l$, and since $l(C)=0$, we have that
there is a definite clause with head $C$ whose body atoms (if it has any)
are all of level $0$ and contained in $I$. Hence condition (Sii) (or (Si))
is satisfied for this clause and $S$ is a self-founded set of $P$ with
respect to $I$.  So assume now that the induction hypothesis holds for all
$B\in B_P$ with $l(B)<\alpha$, and let $A$ be such that $l(A)=\alpha$. We
consider two cases.

(Case i) If $A\in I$, consider the set $S$ of all atoms $B$ with
$l(B)=\alpha$ and $B\in I$. We show that $S$ is a self-founded set of $P$
with respect to $\hitCW_P\!\uparrow\!\alpha$, and this suffices since it implies $A\in
\hitCW_P\!\uparrow\!(\alpha+1)$ by the fact that $A\in S$. First note that $S\subseteq
I$, so $S\cup I\in I_P$. Now let $C\in S$. Then $C\in I$ and $C$ satisfies
condition (Ci) with respect to $I$ and $l$, so there is a clause $A\gets
A_1,\dots,A_n,\lnot B_1,\dots,\lnot B_m$ in $\hitground(P)$ such that
$A_i,\lnot B_j\in I$, $l(A)\geq l(A_i)$, and $l(A)>l(B_j)$ for all $i$ and
$j$. By induction hypothesis we obtain $\lnot B_j\in \hitCW_P\!\uparrow\!\alpha$. If
$l(A_i)<l(A)$ for some $A_i$ then we have $A_i\in\hitCW_P\!\uparrow\!\alpha$, also by
induction hypothesis. If there is no $A_i$ with $l(A_i)=l(A)$, then (Si)
holds, while $l(A_i)=l(A)$ implies $A_i\in S$, so (Sii) holds.

(Case ii) If $\lnot A\in I$, then $A$ satisfies (Fii) with respect to $I$
and $l$. Hence for all clauses $A\gets\texttt{body}$ in $\hitground(P)$ we have that
there is $L\in\texttt{body}$ with $\lnot L\in I$ and $l(L)<\alpha$. Hence for all
these $L$ we have $\lnot L\in\hitCW_P\!\uparrow\!\alpha$ by induction hypothesis, and
consequently for all clauses $A\gets\texttt{body}$ in $\hitground(P)$ we obtain that
$\texttt{body}$ is false in $\hitCW_P\!\uparrow\!\alpha$ which yields $\lnot
A\in\hitCW_P\!\uparrow\!(\alpha+1)$. This establishes fact (2) and concludes the proof
of (i).

(ii) We first introduce some notation. Let
\begin{align*}
L_0 = \emptyset, &\qquad G_0 = B_P,\\
L_{\alpha+1} = \hitCGL_P(G_\alpha), &\qquad
G_{\alpha+1} = \hitCGL_P(L_\alpha)\qquad\text{for any ordinal $\alpha$},\\
L_\alpha = \bigcup_{\beta<\alpha} L_\beta, &\qquad
G_\alpha = \bigcap_{\beta<\alpha} G_\beta\qquad\text{for limit ordinal
  $\alpha$},\\
L_P =\hitlfp(\hitCGL_P^2), &\qquad G_P =\hitgfp(\hitCGL_P^2).
\end{align*}
By transfinite induction, it is easily checked that $L_\alpha\subseteq
L_\beta\subseteq G_\beta\subseteq G_\alpha$ whenever $\alpha\leq\beta$. So
$L_P=\bigcup L_\alpha$ and $G_P=\bigcap G_\alpha$.

Let $M= L_P\cup\lnot (B_P\setminus G_P)$. We intend to apply (i) and first
define an $M$-partial level mapping $l$. We will take as image set of $l$,
pairs $(\alpha,\gamma)$ of ordinals, with the lexicographic ordering. This
can be done without loss of generality since any set of such pairs, under
the lexicographic ordering, is well-ordered, and therefore order-isomorphic
to an ordinal. For $A\in L_P$, let $l(A)$ be the pair $(\alpha,0)$, where
$\alpha$ is the least ordinal such that $A\in L_{\alpha+1}$. For $B\not\in
G_P$, let $l(B)$ be the pair $(\beta,\gamma)$, where $\beta$ is the least
ordinal such that $B\not\in G_{\beta+1}$, and $\gamma$ is least such that
$B\not\in T_{P/L_\beta}\!\downarrow\! \gamma$. It is easily shown that $l$ is
well-defined, and we show next by transfinite induction that $P$ satisfies
(CW) with respect to $M$ and $l$.

Let $A\in L_1=\hitgfp\left(T_{P/B_P}^+\right)$. Since $P/B_P$ contains exactly
all clauses from $\hitground(P)$ which contain no negation, we have that $A$ is
contained in the greatest two-valued model of a definite subprogram of $P$,
namely $P/B_P$. So there must be a definite clause in $\hitground(P)$ with head
$A$ whose corresponding body atoms are also true in $L_1$, which, by
definition of $l$, must have the same level as $A$, hence (Ci) is
satisfied. Now let $\lnot B\in\lnot (B_P\setminus G_P)$ such that $B\in
(B_P\setminus G_1)=B_P\setminus \hitgfp\left(T_{P/\emptyset}^+\right)$. Since
$P/\emptyset$ contains all clauses from $\hitground(P)$ with all negative
literals removed, we obtain that $B$ is not contained in the greatest
two-valued model of the definite program $P/\emptyset$, and (Fii) is
satisfied by Theorem \ref{theo:defgreatest} using a simple induction
argument. 

Assume now that, for some ordinal $\alpha$, we have shown that $A$ satisfies
(CW) with respect to $M$ and $l$ for all $A\in B_P$ with $l(A)< (\alpha,0)$.

Let $A\in L_{\alpha+1}\setminus
L_{\alpha}=\hitgfp\left(T_{P/G_{\alpha}}^+\right)\setminus L_{\alpha}$. Then
$A\in \left(T_{P/G_\alpha}^+\!\downarrow\!\gamma\right)\setminus L_\alpha$ for some
$\gamma$; note that all (negative) literals which were removed by the
Gelfond-Lifschitz transformation from clauses with head $A$ have level less
than $(\alpha,0)$. Then $A$ satisfies (Ci) with respect to $M$ and $l$ by
definition of $l$.

Let $A\in (B_P\setminus G_{\alpha+1})\cap G_\alpha$. Then $A\not\in
\hitgfp\left(T_{P/L_\alpha}^+\right)$ and we conclude again from Theorem
\ref{theo:defgreatest}, using a simple induction argument, that $A$
satisfies (CW) with respect to $M$ and $l$.

This finishes the proof that $P$ satisfies (CW) with respect to $M$ and $l$.
It remains to show that $M$ is greatest with this property.

So assume that $M_1\supset M$ is the greatest model such that $P$ satisfies
(CW) with respect to $M_1$ and some $M_1$-partial level mapping
$l_1$. Assume $L\in M_1\setminus M$ and, without loss of generality, let the
literal $L$ be chosen such that $l_1(L)$ is minimal. We consider two cases.

(Case i) If $L=\lnot A\in M_1\setminus M$ is a negated atom, then by (Fii)
for each clause $A\gets L_1,\dots,L_n$ in $\hitground(P)$ there exists $i$ with
$\lnot L_i\in M_1$ and $l_1(A)>l_1(L_i)$. Hence, $\lnot L_i\in M$ and
consequently for each clause $A\gets\texttt{body}$ in $P/L_P$ we have that some atom
in $\texttt{body}$ is false in $M=L_P\cup\lnot (B_P\setminus G_P)$. But then
$A\not\in\hitCGL_P(L_P)=G_P$, hence $\lnot A\in M$, contradicting $\lnot A\in
M_1\setminus M$.

(Case ii) If $L=A\in M_1\setminus M$ is an atom, then $A\not\in
M=L_P\cup\lnot(B_P\setminus G_P)$ and in particular $A\not\in
L_P=\hitgfp\left(T_{P/G_P}^+\right)$. Hence $A\not\in T_{P/G_P}^+\!\downarrow\!\gamma$
for some $\gamma$, which can be chosen to be least with this property. We
show by induction on $\gamma$ that this leads to a contradiction, to finish
the proof.

If $\gamma=1$, then there is no clause with head $A$ in $P/G_P$, i.e. for
all clauses $A\gets\texttt{body}$ in $\hitground(P)$ we have that $\texttt{body}$ is false in
$M$, hence in $M_1$, which contradicts $A\in M_1$.

Now assume that there is no $B\in M_1\setminus M$ with $B\not\in
T_{P/G_P}^+\!\downarrow\!\delta$ for any $\delta<\gamma$, and let $A\in M_1\setminus
M$ with $A\not\in T_{P/G_P}^+\!\downarrow\!\gamma$, which implies that $\gamma$ is a
successor ordinal. By $A\in M_1$ and (Ci) there must be a clause $A\gets
A_1,\dots,A_n\lnot B_1,\dots,\lnot B_m$ in $\hitground(P)$ with $A_i,\lnot
B_j\in M_1$ for all $i$ and $j$. However, since $A\not\in
T_{P/G_P}^+\!\downarrow\!\gamma$ we obtain that for each $A\gets A_1,\dots,A_n$ in
$P/G_P$, hence for each $A\gets A_1,\dots,A_n,\lnot B_1,\dots,\lnot B_m$ in
$\hitground(P)$ with $\lnot B_1,\dots, \lnot B_m \in \lnot (B_P\setminus
G_P)\subseteq M\subseteq M_1$ there is $A_i$ with $A_i\not\in
T_{P/G_P}^+\!\downarrow\!(\gamma-1)\subseteq M$, and by induction hypothesis we
obtain $A_i\not\in M_1$. So $A_i\in M_1$ and $A_i\not\in M_1$, which is a
contradiction and concludes the proof.
\end{proof}

\section{Related Work}

As the purpose of our paper is to present a coherent unified picture of
different semantics, it is related to the large body of work on relating
semantics and uniform frameworks for semantics of logic programs. For a
subjective selection of the probably most prominent approaches we refer to
the introduction of this paper and also to the extensive discussions in
\cite{HW05}, where the level-mapping approach was introduced and put into
perspective. Two very recent developments, however, appear to be very
closely related to our approach, and we discuss them shortly. They were both
developed independently of our work, and brought to our attention while this
paper was being reviewed.

Loyer, Spyratos and Stamate, in \cite{LSS03}, presented a parametrized
approach to different semantics. It allows to substitute the preference for
falsehood by preference for truth in the stable and well-founded semantics,
but uses entirely different means than presented here. Its purpose is also
different --- while we focus on the strenghtening of the mathematical
foundations for the field, the work in \cite{LSS03} is motivated by the need
to deal with open vs. closed world assumption in some application
settings. The exact relationship between their approach and ours remains to
be worked out.

Denecker, Bruynooghe, Marek, and Ternovska, in \cite{DBM01,DT04lpnmr},
unified different logic programming semantics by identifying them as
transfinite inductive definitions. As the latter can also be analysed using
fixed-point computations via semantic operators, our level-mapping proof
schema as described in \cite{HW05} should be applicable to this inductive
perspective as well. We believe that our approach provides more flexibility
and can be more readily extended to other syntactic and semantic features,
but further work will be needed to substantiate this. On the other hand, the
inductive approach appears to be more intuitively appealing at first sight,
and of more general explanatory value.

\section{Conclusions and Further Work}\label{sec:conc}

We have displayed a coherent picture of different semantics for normal logic
programs. We have added to well-known results new ones which complete the
formerly incomplete picture of relationships. The richness of theory and
relationships turns out to be very appealing and satisfactory. 

As noted already in the introduction, we did not intend to provide new
semantics for practical purposes. We rather wanted to focus on the deepening
of the theoretical insights into the relations between different semantics,
by painting a coherent and complete picture of their dependencies and
interconnections. Nevertheless, our new semantics stands well in the
tradition of the original motivation of non-monotonic reasoning research:
our semantics is defined by making a selection of the (classical) models of
a program, understood as first-order logical formulae. We do not claim that
the this line of motivation necessarily carries much further --- as
repeatedly stated, our purpose is formal, and foundational.

From a mathematical perspective one expects major notions in a field to be
strongly and cleanly interconnected, and it is fair to say that this is the
case for declarative semantics for normal logic programs, as our exhibition
shows.  We would also like to stress that the results presented in this
paper are far from straightforward. Intuitively, replacing least fixed
points by greatest fixed points appears to be unproblematic, but this is
only true on the general conceptual level, and far from obvious, or easy to
achieve, formally. The details of the constructions and proofs are indeed
involved and not incremental, which is particularly obvious by the proof
details for Theorem
\ref{theo:main}. The fact that a symmetric picture such as the one presented
here can be established at all is strongly supportive of the position that
major established notions in logic programming are not only intuitively
appealing --- this is well-known as intuition was one of the driving forces
in the field --- but also formally satisfactory.

For normal logic programs, we have obtained a uniform perspective on
different semantics.  The situation becomes much more difficult when
discussing extensions of the logic programming paradigm like disjunctive
\cite{Wan01}, quantitative \cite{Mat00}, or dynamic \cite{Lei03} logic
programming. For many of these extensions it is as yet to be determined what
the best ways of providing declarative semantics for these frameworks are,
and the lack of interconnections between the different proposals in the
literature provides an argument for the case that no satisfactory answers
have yet been found.

We believe that successful proposals for extensions will have to exhibit
similar interrelationships as observed for normal programs. How, and if,
this can be achieved, however, is as yet rather uncertain. Formal studies
like the one in this paper may help in designing satisfactory semantics, but
a discussion of this is outside the scope of our exhibition, and will be
pursued elsewhere.

\end{document}